\documentclass[letterpaper, 10pt, journal, twoside]{IEEEtran}

\usepackage{cite}
\usepackage[bookmarks=true, colorlinks=false,
    linkcolor=black,
    filecolor=mrobota,
    urlcolor=cyan]{hyperref}
\usepackage{amsmath}
\usepackage{algorithmicx}
\usepackage{algpseudocode}
\usepackage{adjustbox}

\usepackage{amsfonts, amssymb}
\usepackage{algorithm}
\usepackage{soul}
\usepackage{tabularx}
\usepackage[dvipsnames]{xcolor}
\usepackage[caption=false,font=normalsize,labelfont=sf,textfont=sf]{subfig}
\usepackage{adjustbox}
\usepackage{booktabs}
\usepackage{graphicx}
\usepackage{csvsimple}
\usepackage{amsthm}
\usepackage{tikz}
\usepackage{pgf}
\usepackage{import}
\usepackage{pgfplots}
\usepackage{pgfplotstable}
\usepgfplotslibrary{fillbetween}
\usepackage{multirow}
\usepackage{siunitx}
\usepackage{svg}

\usepackage[utf8]{inputenc}

\newtheorem*{problem}{Problem}

\usepackage{stackengine}
\usepackage{color}
\usepackage{todonotes}
\usepackage{amssymb}
\usepackage{mathtools}

\usepackage{booktabs}   %
\usepackage{array}      %
\usepackage{makecell}   %
\usepackage{pifont}     %

\usepackage{orcidlink}
\newcommand{\noboxorcid}[1]{%
  {\hypersetup{pdfborder={0 0 0}}\href{https://orcid.org/#1}{\orcidlink{#1}}}%
}

\usepackage{comment}
\title{Energy-Efficient Multi-Robot Coverage Path Planning of Non-Convex Regions of Interests}
\author{{Sourav Raxit$^{1}$\noboxorcid{0000-0003-1196-2435}}, {Jose Fuentes$^{2}$\noboxorcid{0000-0002-6477-5820}}, {Paulo Padrao$^{3}$\noboxorcid{0000-0003-3966-0279}}, {Abdullah Al Redwan Newaz$^{1^*}$\noboxorcid{0000-0003-1140-8119}}, {Md Tamjidul Hoque$^{1}$\noboxorcid{0000-0002-0110-2194}}, {Mark Kulp$^{1}$\noboxorcid{0000-0002-9724-6917}}, and {Leonardo Bobadilla$^{2}$\noboxorcid{0000-0003-2097-2432}}
\thanks{Manuscript received: December 31 2025; Revised: February 23, 2026; Accepted April 21 2026. This paper was recommended for publication by editor Soon-Jo Chung upon evaluation of the Reviewers' comments.
This work is supported in part by the U.S. EPA grant BR-02F47801-5010M, NSF grants 2118329, IIS-2024733 and IIS-2331908, the ONR grant N00014-23-1-2789, theU.S. DoD grant 78170-RT-REP, and by the Army Research Laboratories under contract W911NF1920243. This is contribution \#2139 from the Institute of Environment at FIU.}
\thanks{$^*$Corresponding Author :Abdullah Al Redwan Newaz. $^{1}$ S. Raxit, A. A. R. Newaz, and Md Tamjidul Hoque are with the Department of Computer Science, and M. Kulp is with the Department of Earth and Environmental Sciences, University of New Orleans, New Orleans, LA 70148, USA (email: \{sraxit, aredwann, thoque, mkulp\}@uno.edu).
$^{2}$ J. Fuentes, and L. Bobadilla are with the School of Computing and Information Sciences, Florida International University, Miami, FL 33199, USA (email:
        \{jfuen099@, bobadilla@cs.\}fiu.edu).
$^{3}$ P. Padrao is with  Providence College, Department of Mathematics \& Computer Science, Providence,   RI 02918  (email:ppadraol@providence.edu).
        }
    \thanks{Digital Object Identifier (DOI): see top of this page.}
}
\begin{document}
    \maketitle
    \begin{abstract}

    This letter presents an energy-efficient multi-robot coverage path planning (MRCPP) framework for large, nonconvex Regions of Interest (ROI) containing obstacles and no-fly zones (NFZ). Existing minimum-energy coverage planning algorithms utilize meta-heuristic boustrophedon workspace decomposition. Therefore, even with minimum energy objectives and energy consumption constraints, they cannot achieve optimal energy efficiency. Moreover, most existing frameworks support only a single type of robotic platform. MRCPP overcomes these limitations by: generating globally-informed swath generation, creating parallel sweeping paths with minimal turns, calculating safety buffers to ensure safe turning clearance, using an efficient mTSP solver to balance workloads and minimize mission time, and connecting disjoint segments via a modified visibility graph that tracks heading angles while maintaining transitions within safe regions.
    The efficacy of the proposed MRCPP framework is demonstrated through real-world experiments involving autonomous aerial vehicles (AAVs) and autonomous surface vehicles (ASVs).
    Evaluations demonstrate that the proposed MRCPP consistently outperforms state-of-the-art planners, reducing average total energy consumption by 3\% to 40\% for a team of 3 robots and computation time by an order of magnitude, while maintaining balanced workload distribution and strong scalability across increasing fleet sizes.
    The MRCPP framework is released as an open-source package and videos of real-world and simulated experiments are available at \url{https://mrc-pp.github.io/}.

\end{abstract}


\section{Introduction}
\IEEEPARstart{E}{nergy-efficient} Coverage Path Planning (CPP) is crucial for autonomous robots to systematically traverse a Region of Interest (ROI), supporting applications in precision agriculture~\cite{koutras2020autonomous}, environmental monitoring~\cite{kapoutsis2019distributed}, infrastructure inspection~\cite{shakhatreh2019unmanned}, and search-and-rescue~\cite{erdos2013experimental}. 
Energy consumption can be minimized either explicitly, by incorporating battery energy constraints into the planning process, or implicitly, by optimizing surrogate metrics closely related to energy usage, such as total route length or the number of turns~\cite{datsko2024energy}.

Recent CPP methods~\cite{datsko2024energy, Mier_Fields2Cover_An_open-source_2023, apostolidis2022cooperative, shah2022large}, either utilize boustrophedon decomposition~\cite{choset2000coverage} or employ grid-based sweeping~\cite{bahnemann2021revisiting}. These approaches perform effectively in convex, obstacle-free domains but often struggle in non-convex ROIs featuring irregular boundaries and internal exclusion zones~\cite{galceran2013survey}. However, meta-heuristic decomposition results in fragmented swaths, excessive turns, and redundant overlaps, thereby increasing energy consumption and mission time. Compared to single-robot CPP, multi-robot CPP (MRCPP) frameworks significantly accelerate coverage by enabling parallel execution of robots. However, they face substantial challenges in simultaneously satisfying three tightly interdependent requirements: globally-informed swath generation, obstacle-aware path generation, and balanced task allocation.

Existing MRCPP methods inadequately minimize turns and ensure path continuity due to their dependence on meta-heuristic decomposition. Partition-based techniques (e.g., DARP+MST~\cite{kapoutsis2017darp,apostolidis2022cooperative}) ignore sweep orientation, leading to inconsistent headings and excessive turns. Cell-based GTSP solvers (e.g., POPCORN+SALT~\cite{shah2022large,fischetti1995symmetric}) generate disjointed paths with frequent reversals and scale poorly in complex domains due to NP-hard complexity. Energy-aware methods like EAMCMP~\cite{datsko2024energy} model dynamics but lack robust obstacle-aware load balancing and remain suboptimal due to boustrophedon decomposition. An energy-efficient, computationally lightweight framework that explicitly optimizes turns and workload balance is therefore required.






\begin{figure}[t]
\centering


\includegraphics[
        width=0.47\columnwidth,height=3cm,
        trim=40 0 40 0, clip
    ]{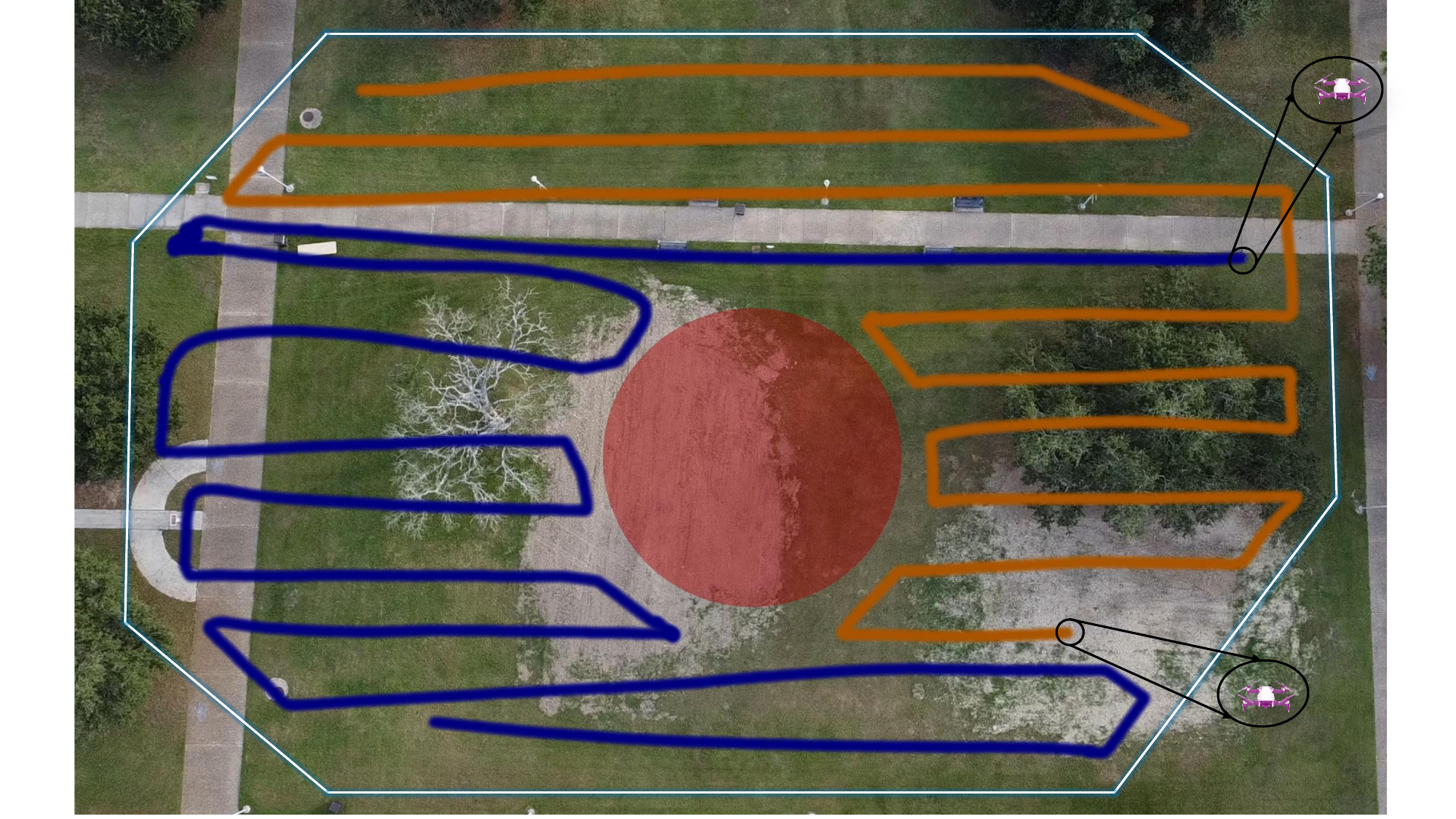}
\hfil
   \includegraphics[
        width=0.47\columnwidth,height=3cm,
        trim=20 20 00 13, clip
    ]{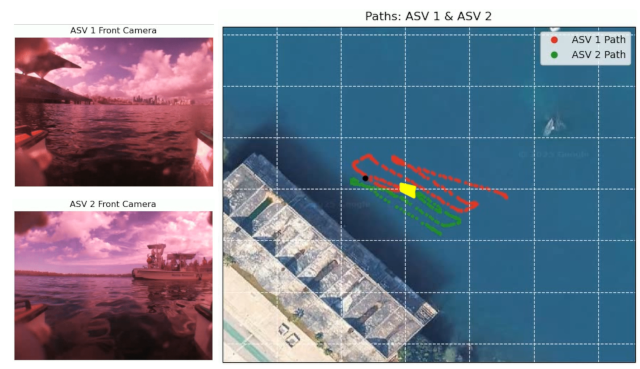}
\caption{Real-world multi-robot coverage experiments. (a)~Two AAVs cover a polygonal ROI (white) while avoiding a no-fly zone (red circle) in the left figure. (b)~Two ASVs cover an aquatic region while avoiding a polygonal obstacle (yellow) in the right figure.}
\label{fig:field_exp}

\end{figure}

In this study, we propose an energy-efficient MRCPP framework for large, non-convex regions of interest containing obstacles and no-fly zones.
We first compute the minimum-area enclosing rectangle using a rotating-calipers algorithm~\cite{toussaint1983solving} to determine a globally-informed sweep orientation. We then generate parallel coverage swaths with minimal turns, clip them to the obstacle-free space, and order them along the minor axis to promote spatial continuity and reduce inter-swath transitions.
Next, we calculate headland buffers that shrink the outer ROI boundary and expand internal exclusion zones~\cite{Mier_Fields2Cover_An_open-source_2023}, creating a safe operable area that ensures adequate turning clearance and prevents boundary violations.
We assign the swaths to multiple robots using an efficient multiple Traveling Salesman Problem (mTSP) solver~\cite{helsgaun2017extension}, which balances individual workloads and minimizes total mission time.
Finally, we connect disjoint swath segments into contiguous sweeping paths with a modified visibility graph (VG) algorithm~\cite{lavalle2006planning}  augmented with sampled boundary points and a heading-driven traversal rule that keeps all transitions within the buffered regions.

 \textbf{Contributions:}
(i) An \textbf{orientation-optimized swath generation algorithm} that minimizes turns by aligning coverage with the ROI's principal axis while maintaining obstacle and safety buffer clearance (Section~\ref{section:swath} ).
(ii) A \textbf{workload-balanced mTSP formulation} for multi-robot swath allocation, which balances individual workloads and minimizes overall mission time (Section~\ref{section:multi_robot}).
(iii) A \textbf{collision-free, contiguous sweeping path generation} algorithm based on a modified VG algorithm that augments the node set with sampled boundary points and uses heading-driven entry/exit selection to maintain feasible transitions (Section~\ref{section:multi_robot}).

\section{Related Work}
\label{sec:related_work}

Early research in CPP focused on workspace decomposition using metaheuristic approaches, such as boustrophedon decomposition~\cite{choset1998coverage,choset2000coverage}, to handle regions with obstacles while reducing the number of subcells generated by an exact cellular decomposition method~\cite{bahnemann2021revisiting}. Although path optimization of these methods involves spiral~\cite{di2016coverage} or graph~\cite{shah2020multidrone} representations over decomposed regions to cluster coverage tasks, they often struggle with fragmented paths and increased turning effort in highly non-convex or cluttered regions. 

To address these limitations, researchers have reformulated CPP as a routing optimization problem similar to the Generalized Traveling Salesman Problem (GTSP), where combining decomposition with GTSP-based sequencing can minimize total flight time~\cite{bahnemann2021revisiting}. However, optimizing only for distance overlooks a critical factor: energy consumption is heavily influenced by velocity changes, acceleration, and turns~\cite{bauersfeld2022range}. Energy-aware planners~\cite{di2016coverage,datsko2024energy} incorporate detailed power models and solve clustered TSP variants~\cite{nekovavr2021multi}, yet most classical and decomposition-based methods still use fixed swath orientations rather than optimizing sweep direction, which significantly affects both turn count and energy use.

Minimizing turns has gained attention as a direct path to reducing energy and mission time, since every heading change forces the vehicle to slow down, rotate, and accelerate again~\cite{datsko2024energy}. While several approaches explicitly target turn reduction~\cite{vandermeulen2019turn,stefanopoulou2024improving}, they typically work within predetermined cell boundaries or fixed grid layouts. For instance, in~\cite{stefanopoulou2024improving} added turn awareness to DARP, but the underlying partition-first structure limits the ability to globally coordinate sweep alignment across sub-regions.

Unlike single-robot CPP, where path length and turn count can be optimized over a single contiguous route, multi-robot CPP requires jointly solving region partitioning, sweep alignment, and workload balance. These challenges are tightly coupled and cannot be addressed by simply replicating CPP methods per robot. Early solutions~\cite{apostolidis2022cooperative,barrientos2011aerial,puig2011new} resulted in uneven task distribution across irregular terrain. Voronoi-based methods~\cite{breitenmoser2010voronoi} provide elegant mathematical guarantees but were designed for static sensor placement rather than path planning, ignoring factors like starting positions and travel distances. DARP~\cite{kapoutsis2017darp} explicitly considers these elements along with obstacles and workload balance. However, both DARP and its recent improvements~\cite{stefanopoulou2024improving} partition the space first and plan sweeps second, which can lead to disconnected regions with conflicting sweep directions~\cite{almadhoun2019survey}. Similarly, capability-aware decomposition frameworks~\cite{gray2025multi} 
partition the ROI based on UAV sensing and flight characteristics before 
generating sweep paths and optimizing trajectories, which may still result 
in suboptimal global alignment of coverage directions.

SMT-based decomposition to coordinate multiple AAVs is used in large-scale deployments~\cite{shah2020multidrone,shah2022large}. However, their main limitation is enforcing independence among sub-areas, which prevents globally aligned swaths and can create fragmented coverage patterns. 
Spanning-tree approaches~\cite{agmon2006constructing} maintain coverage guarantees with provable bounds on solution quality, but they don't explicitly optimize for turns or continuous swath structures.

Integrated frameworks~\cite{kapoutsis2017darp,stefanopoulou2024improving} combine DARP-based partitioning with spanning-tree coverage, but their partition-first structure ignores sweep orientation, leading to inconsistent headings and excessive turns at partition boundaries. Energy-aware methods~\cite{di2016coverage,datsko2024energy} incorporate aerodynamic power models but remain suboptimal due to the same decomposition bottleneck. In contrast, MRCPP first determines a near-optimal  sweep orientation to minimize turns, then balances workload across robots via mTSP, and ensures obstacle-aware transitions through an augmented visibility graph, addressing limitations that no existing method resolves jointly.

    \section{Problem Formulation}

\label{sec:problem_formulation}

Consider a team of $N_R$ mobile robots tasked with covering a bounded, possibly nonconvex ROI $P \subset \mathbb{R}^2$. The ROI may contain $N_{\mathrm{obs}}$ exclusion zones such as obstacles, no-fly regions, or restricted areas, modeled as polygonal sets $o_i \subset \mathbb{R}^2$ for $i = 1,\dots, N_{\mathrm{obs}}$.
For simplicity, all exclusion zones are either given as polygons or approximated as such to support polygon-based decomposition. 
The free space $F$ available for coverage is then computed by subtracting obstacle regions from $P$, such that  $F = P \setminus \bigcup_{i=1}^{N_{\mathrm{obs}}} o_i$.

Each robot has a coverage width $w$ that defines both the spacing between adjacent parallel swaths and the distance on either side of a swath line within which points are considered covered. A swath refers to the path segment a robot follows when traversing the field.
Formally, let $S$ be the set of all $N_S$ swath segments such that $S = \{S_1, S_2, \dots, S_{N_S}\}$. Each swath segment $S_m \in S$ is defined by its endpoints via $S_m = [a_m, b_m]$, with geometric length $\ell_m = \|b_m - a_m\|$.

To achieve energy efficiency, swaths should be generated in a way that minimizes the number of turns and the total path length~\cite{datsko2024energy}.
Given a set of swaths $S$ produced by a minimum-turn algorithm, the objective of the multirobot coverage planner is to allocate a subset of swaths $S_R \subset S$ to the $J^{th}$ robot such that every swath segment in $S$ is visited by exactly one robot following a continuous, collision-free sequence. The robots start from given depot locations (or algorithm may assign optimal initial locations) and must traverse their assigned swaths in some order. Between the end of one swath and the start of the next assigned swath, each robot follows a shortest collision-free transition path that remains inside a slightly enlarged feasible space $F'$ (e.g., $F$ offset inward by the $w$ for safe turning). \color{black} Let $d_{\text{trans}}$ denote the length of such a transition path. Therefore, the total distance traveled by robot $r$ can be computed by  
\begin{equation}
    L_r = \sum_{S_m \in \mathcal{T}_r} \ell_m + \sum_{\text{transitions of } r} d_{\text{trans}},
\end{equation}

Assuming all robots operate under a uniform speed model, the overall mission completion time is dictated by the robot with the maximum total executed path length.
\begin{problem}
    Find a partition of the swath set $S$ into $N_R$ subsets $\mathcal{T}_1, \dots, \mathcal{T}_{N_R}$ and, for each robot $r$, a collision-free ordering of its assigned swaths together with corresponding transition paths, such that
\begin{equation}
{\min_{\Pi} \; \sum_{r=1}^{N_R} L_r, \quad \text{s.t.} \quad |\mathcal{T}_r| \geq \left\lfloor \frac{N_S}{N_R} \right\rfloor, \;\; \forall\, r},
\label{eq:mainObjective}
\end{equation}
where $\Pi$ denotes the set of all valid assignments over coverage paths. The minimum tour size constraint ensures that each robot is assigned at least $\lfloor N_S / N_R \rfloor$ swaths, preventing degenerate allocations and promoting approximately balanced workloads. Together with minimizing total path length, this indirectly minimizes the makespan by discouraging unbalanced partitions.
\end{problem}

    \section{Swath Generation For Nonconvex Fields}
    \label{section:swath}
    \begin{figure*}[t]
    \centering
    \includegraphics[width=\textwidth, trim = 2cm 0.6cm 2cm 0.6cm, clip]{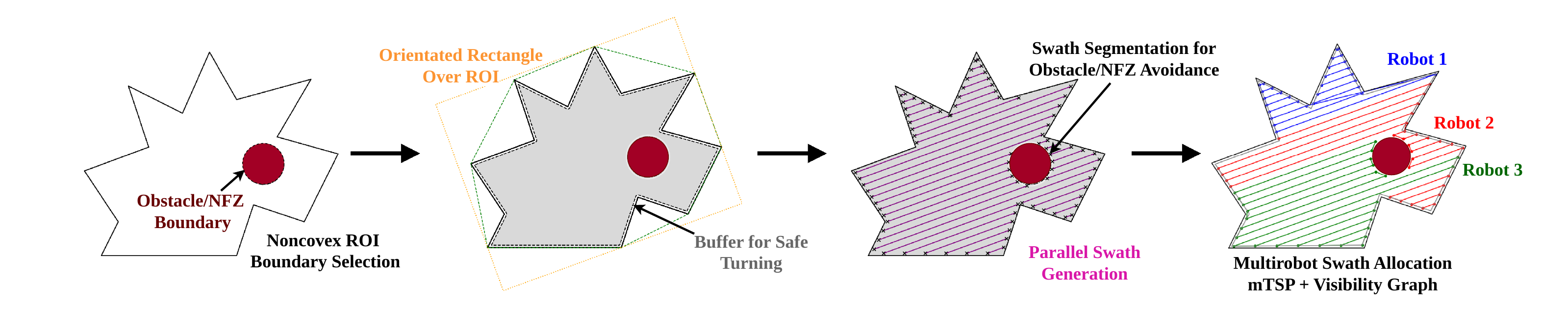}
\vspace{-12 pt}
\caption{\small{Illustration of the MRCPP-based multi-robot coverage workflow. Starting from an input workspace with an internal NFZ, the method computes a headland-adjusted ROI and a minimum-area enclosing rectangle to determine the sweep orientation, generates parallel sweep lines over the free space, and finally allocates the resulting swaths to three robots using mTSP.}}
\label{fig:coverage_flow}
\vspace{-10pt}
\end{figure*}

    When a team of cooperative robots needs to cover a nonconvex ROI, the swath-generation process must ensure that the coverage lines (swaths) are well-aligned with the field’s geometry while minimizing overlaps, gaps, and unnecessary maneuvers. Here, we introduce a minimum-turn swath-generation algorithm that identifies an oriented rectangle for a non-convex ROI and then generates swaths inside the ROI parallel to the long axis of the oriented rectangle.

\subsection{Oriented Rectangle over a Non-Convex ROI}
\label{subsec:min-rect}
Several strategies exist for determining sweep orientation, and while no single method optimally handles all non-convex scenarios, their performance differences are generally comparable. 
\paragraph{Minimum-Area Rectangle (MAR)}
Given a polygonal region of interest $P \subset \mathbb{R}^2$ with $n_P$ vertices, 
the objective is to determine the minimum-area rectangle enclosing $P$ among all possible 
orientations. Since the minimum bounding rectangle must be tangent to the outermost boundary 
of the shape, its computation depends solely on the convex hull $H = \mathrm{ConvHull}(P)$.
We employ the rotating calipers method \cite{toussaint1983solving} to identify the orientation 
of the rectangle that yields minimal area. This algorithm evaluates each edge of the convex hull 
as a potential alignment for one side of the rectangle, computes the corresponding bounding 
rectangle dimensions in that orientation, and runs in linear time with respect to the number of 
hull vertices $n_H$. The resulting oriented rectangle $R_{\min}$ is characterized by its principal 
orientation vectors $(\hat{\mathbf{u}}, \hat{\mathbf{v}})$ and its dimensions $(w_{\min}, h_{\min})$, 
which bound both the convex hull and the original region $P$.
\paragraph{Exhaustive angle search}
This strategy evaluates a discretized set of candidate sweep angles over a prescribed range and selects the one that optimizes a chosen criterion, such as the number of swaths, overlap, or estimated traversal cost.
\paragraph{Principal Component Analysis (PCA)}
This strategy determines the dominant geometric direction of the ROI from the principal axis of its vertex distribution and uses that dominant axis as the sweep orientation.
\paragraph{Minimum-width}
This strategy selects the orientation that minimizes the width of the ROI measured orthogonally to the sweep direction, which can reduce the number of required back-and-forth coverage passes.

    \subsection{Minimum-turn Swath Generation}
\label{subsec:swath_gen}

Once a minimum rotating rectangle is found, we then partition the ROI into parallel coverage lines aligned with the major axis $\mathbf{u}$. Next, we project all vertices of $P$ onto the perpendicular axis $\mathbf{v}$ to maximize straight-line traversal while reducing the turning effort as:
\begin{equation}    
\eta_{\min} = \min_{\mathbf{p} \in V_P} (\mathbf{p} \cdot \mathbf{v}),
\;\;\eta_{\max} = \max_{\mathbf{p} \in V_P} (\mathbf{p} \cdot \mathbf{v}),
\label{eq:eta_min}
\end{equation}

where $V_P$ denotes the set of all vertices of the polygonal region $P$ and the total projection span is $\Delta\eta = \eta_{\max} - \eta_{\min}$.

Given a swath width $w$, the number of required swaths is $n_s = \left\lceil \frac{\Delta\eta}{w} \right\rceil.$
The $k$-th swath center offset is $c_k = \eta_{\min} + \frac{w(k+1)}{2} ,
\qquad k = 1,\dots,n_s,$
where $k$ indexes the swaths and $n_s$ is the total number of swaths computed from the projection span $\Delta\eta$ and the corresponding infinite swath line is given by $x(t) = c_k \mathbf{v} + t \mathbf{u}$ for $t \in \mathbb{R}$.

To obtain valid in-field segments, each swath line is clipped against the polygonal boundary of $P$. Let $E_\mu = [\mathbf{a}_\mu,\mathbf{b}_\mu]$, $\mu = 1,\dots,n_e$, denote the edges of the polygonal boundary of $P$, where $n_e$ is the total number of boundary edges. Each edge is parameterized as $e_\mu(s) = \mathbf{a}_\mu + s(\mathbf{b}_\mu - \mathbf{a}_\mu)$ for $s \in [0,1]$. 
Intersecting the swath line with each edge yields
\begin{equation}   
t_\mu = 
\frac{(\mathbf{a}_\mu - c_k \mathbf{v})\times(\mathbf{b}_\mu - \mathbf{a}_\mu)}
     {\mathbf{u} \times (\mathbf{b}_\mu - \mathbf{a}_\mu)}, \;\;
     s_\mu =
\frac{(\mathbf{a}_\mu - c_k \mathbf{v})\times \mathbf{u}}
     {\mathbf{u} \times (\mathbf{b}_\mu - \mathbf{a}_\mu)}.
\label{eq:tmu}
\end{equation}

All intersections satisfying $s_\mu \in [0,1]$ are retained and sorted into the ordered points $q_1,\dots,q_{m_k}$. The candidate swath segments are
\begin{equation}
S_{k,\nu} = [q_\nu,\, q_{\nu+1}], 
\qquad \nu = 1,\dots,m_k - 1.
\label{eq:skv}
\end{equation}

A segment is valid if its midpoint lies inside the ROI. Interior membership is determined using the even--odd rule, a point-in-polygon test in which a ray cast from the midpoint intersects the polygon edges $E_\mu$ an odd number of times if the point lies inside 
$P$ and an even number of times otherwise. The complete set of minimum-turn swaths is written compactly as $S = \bigcup_{k=1}^{n_s} \{ S_{k,\nu} \subset P \}$. However, in practice, swaths are generated inside the buffered feasible region $P^\prime$ (Section~\ref{sec:buffer}) rather than $P$, ensuring that all segments lie within the safe operable area with adequate turning clearance.

    \section{Multirobot Coverage Path Planning}
    \label{section:multi_robot}
    Given the set of minimum-turn swaths $S$, the objective is to allocate swaths
among robots and construct continuous, collision-free coverage paths entirely
contained within the buffered free space.

\subsection{Buffer for Safe Turning and Obstacle Avoidance}\label{sec:buffer}

We dilate the ROI boundary and all exclusion zones to ensure safe turning clearance and obstacle avoidance. In agricultural robotics, the term \textit{headland} refers to the strip reserved at field edges for vehicle turning~\cite{Mier_Fields2Cover_An_open-source_2023}; we adopt the same concept here. The dilation is performed by offsetting a headland width $h$ using the Minkowski operation.  Let $B_h = \{ x \in \mathbb{R}^2 : \|x\| \le h \}$ denote the closed disk of radius $h$. The inward offset of the ROI is computed as the Minkowski difference $P' = P \ominus B_h,$
 where the Minkowski difference is defined as $A \ominus B = \{a - b \mid a \in A,\, b \in B\}$~\cite{lavalle2006planning}.
while each exclusion zone is expanded outward using the Minkowski sum
\begin{equation}
o_i' = o_i \oplus B_h,
\qquad i = 1,\dots,N_{\mathrm{obs}}.
\label{eq:offset_out}
\end{equation}

Therefore, we can compute the feasible coverage space as:
\begin{equation}
F' = P' \setminus \bigcup_{i=1}^{N_{\mathrm{obs}}} o_i'.
\label{eq:Fprime}
\end{equation}

These offset boundaries also provide nodes for constructing the visibility graph used in the subsequent transition-planning stage. The ROI/NFZ representation, headland-adjusted feasible region, sweep orientation, and resulting parallel swaths are illustrated in Fig.~\ref{fig:coverage_flow}.

\subsection{Multi-Robot Swath Allocation}
\label{subsec:allocation}


\begin{figure*}[h!]
\centering
\resizebox{\textwidth}{!}{%
\subfloat[Cape\label{subfig:cape}]{%
    \includegraphics[height=7.0cm]{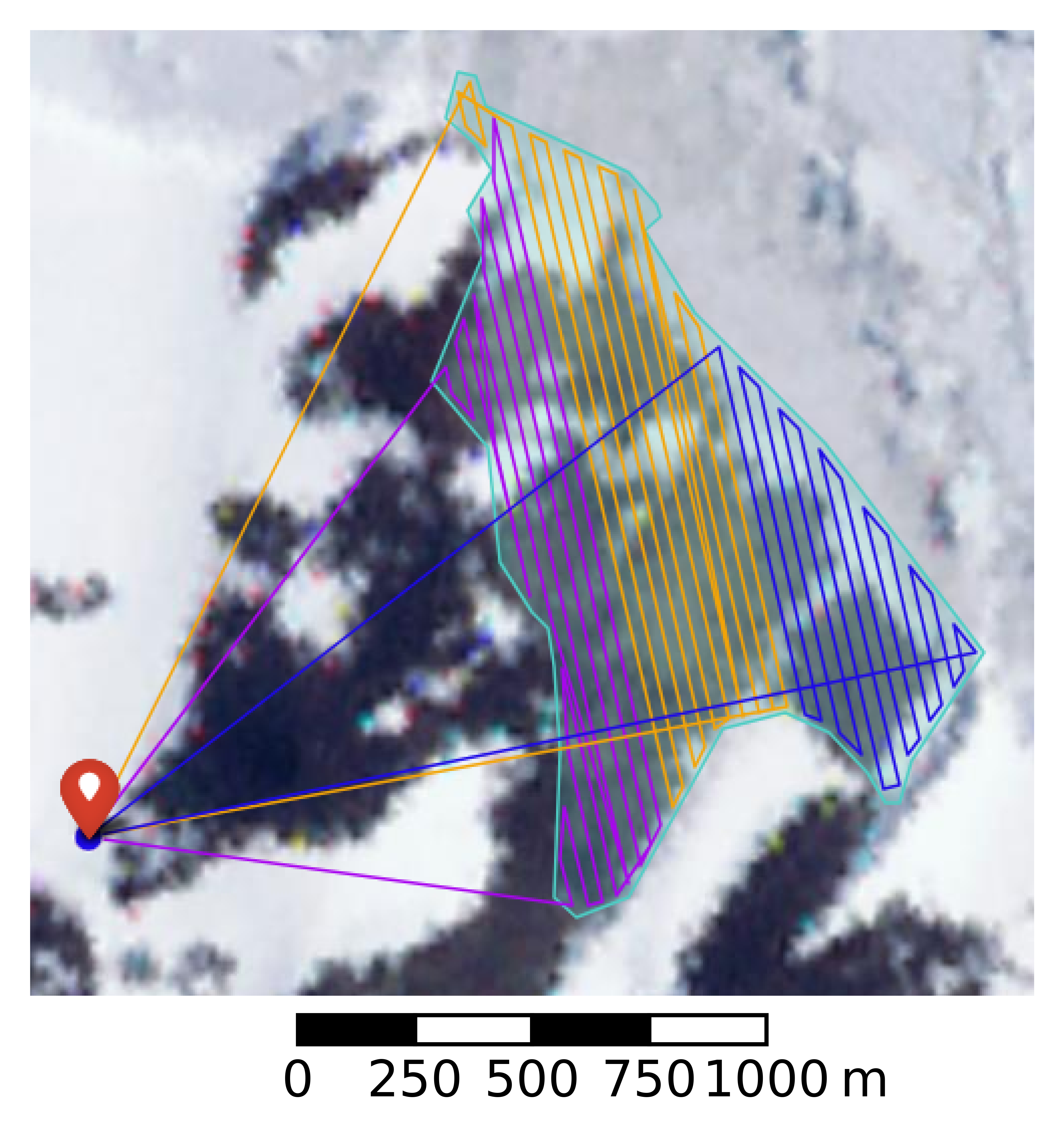}}%
\subfloat[Complex\label{subfig:complex}]{%
    \includegraphics[height=7.0cm]{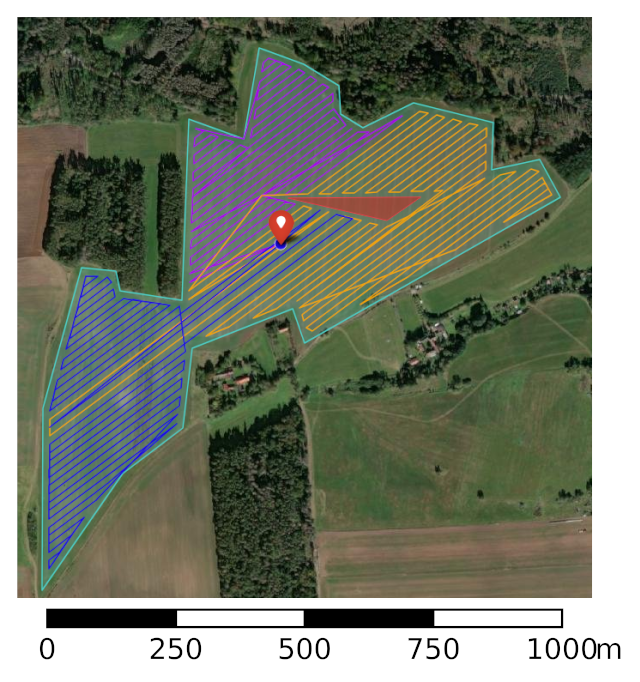}}%
\subfloat[Wetland\label{subfig:island}]{%
    \includegraphics[height=7.0cm]{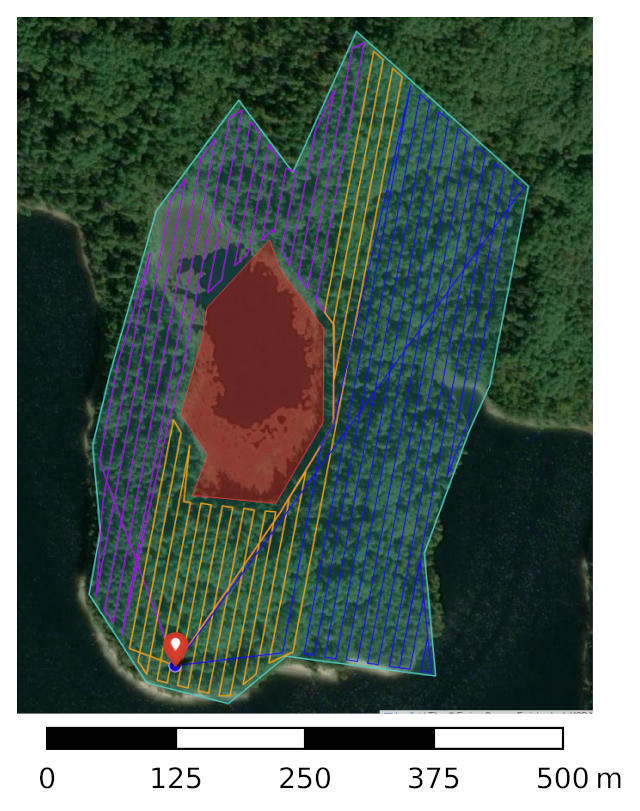}}%
\subfloat[Simple\label{subfig:simple}]{%
    \includegraphics[height=7.0cm]{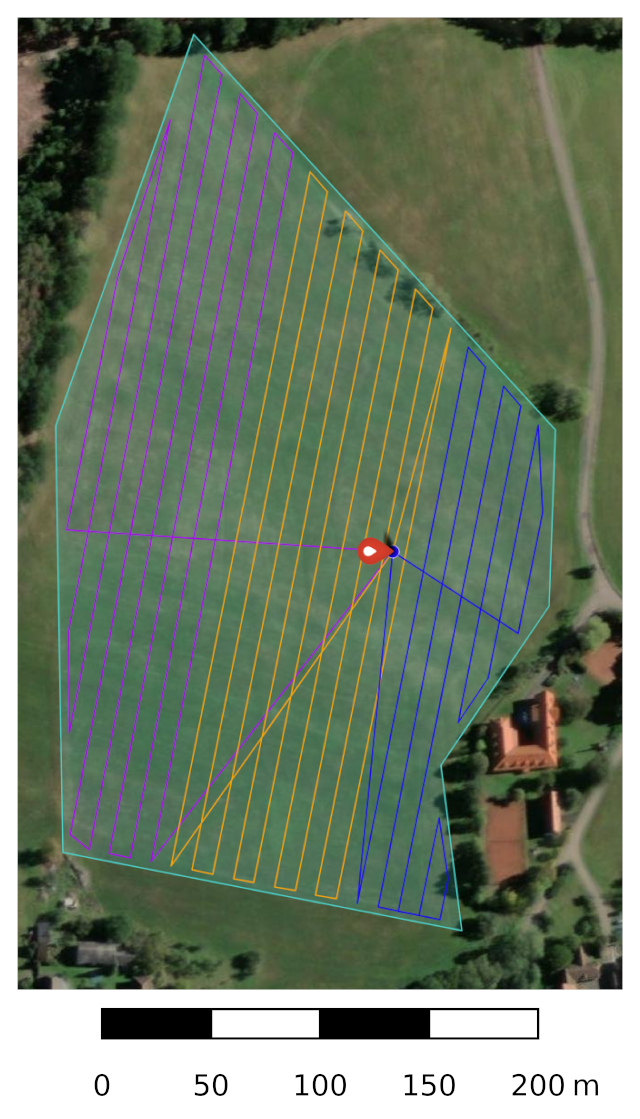}}%
\subfloat[Rectangle\label{subfig:rect}]{%
    \includegraphics[height=7.0cm]{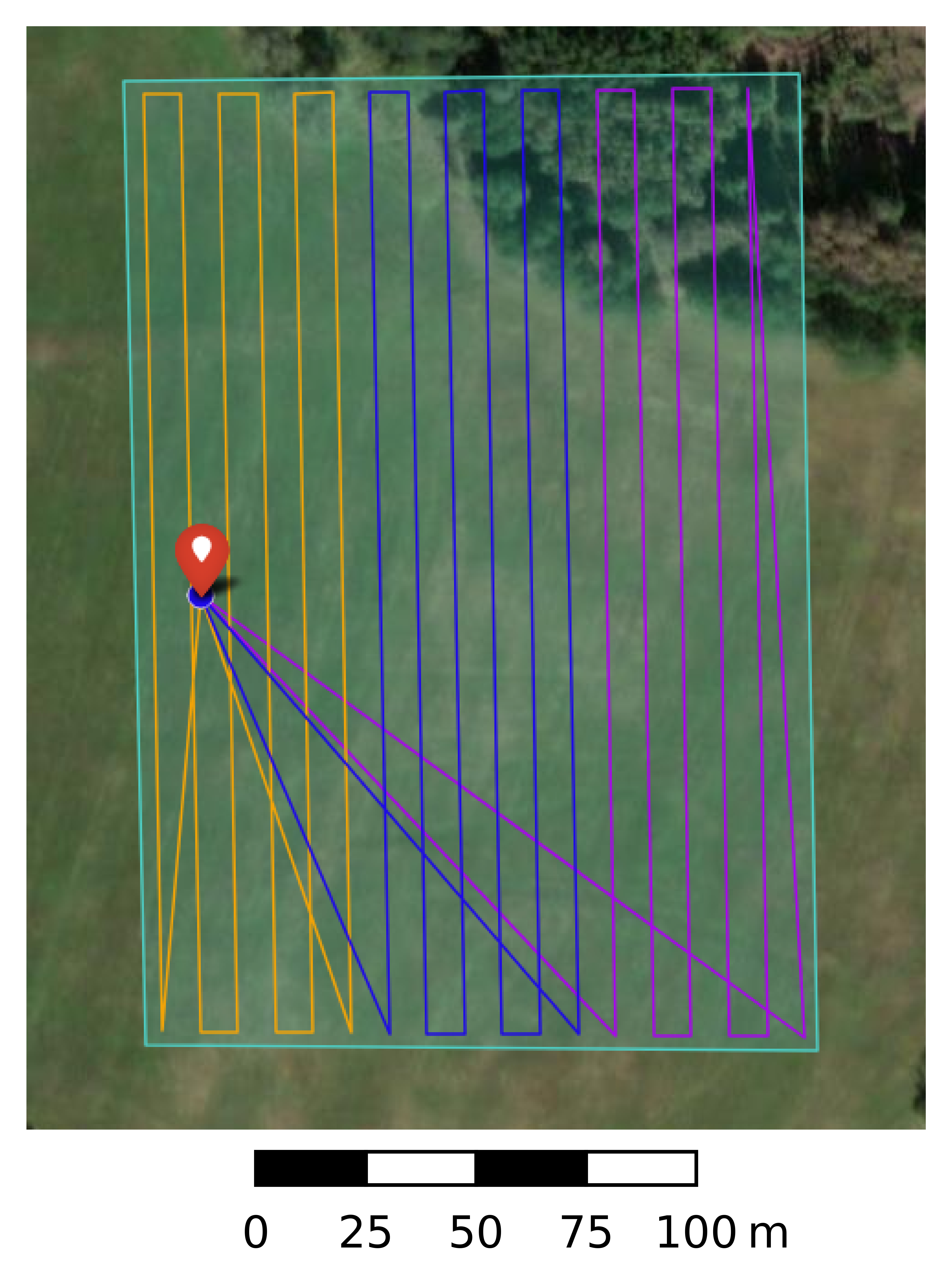}}%
}
\vspace{-10 pt}
\caption{Representative benchmark environments with coverage paths for three AAVs in yellow, violet, and blue. The AAV paths are generated by our MRCPP method within the given polygons. In (b) Complex and (c) Wetland, No-Fly Zones (NFZs) are shown as shaded red polygons. Scale bars are shown in meters.}
\vspace{-12 pt}
\label{fig:envs}
\end{figure*}

The multi-robot swath allocation module distributes the set of minimum-turn swaths among the robots to balance the overall mission workload and minimize the total mission completion time. For this purpose, each swath $S_m = [a_m, b_m]$ is assigned a centroid as:
$c_m = \frac{a_m + b_m}{2},$
which serves as a representative geometric point for the swath and is used to determine its relative position within the region. The swaths are then ordered according to the projection $z_m = c_m \cdot v$, which sorts the swaths along the sweep normal direction and enforces a consistent global traversal order across the field. Depending on the robot heading angle, it enters the swath of its two endpoints, $E_m = \{a_m, b_m\}$. 
 To estimate the true travel cost between two swath endpoints, we consider feasible transitions inside $F'$. If a direct segment between two endpoints
is obstructed, the shortest collision-free path is computed using a visibility
graph constructed over the boundary of $F'$ and obstacle polygons.
\begin{equation}
d_{F'}(x,y) =
\begin{cases}
\|x-y\|, & x,y \in F', \\
\text{VG shortest path}, & \text{otherwise}.
\end{cases}\label{eq:distance}
\end{equation}
As shown in Eqn.~\eqref{eq:distance}, if a direct straight-line segment lies entirely inside the feasible region, the Euclidean distance is used; otherwise, the shortest collision-free path obtained from the visibility graph is substituted.

Next, the transition cost between two swaths is computed as:
\begin{equation}
c_{m,m'} =
\min_{x \in E_m,\; y \in E_{m'}} d_{F'}(x,y).
\end{equation}
This cost selects the minimum feasible transition distance over all possible entry–exit endpoint combinations of the two swaths.  Note that the mTSP formulation considers all endpoint combinations to minimize transition cost, but does not encode swath traversal direction as a two-endpoint state variable; the actual heading assignment is resolved post~hoc during sweep path generation (Section~\ref{subsec:sweep-path}). Because $c_{m,m'}$ is a lower bound on the realized transition cost, it serves as an optimistic proxy; however, for spatially contiguous, parallel swaths---which the solver preferentially groups---the boustrophedon pattern selects the same endpoint pair that achieves the minimum, so the bound is tight in the common case.

Finally, we assign swaths to multiple robots by minimizing the total travel cost across the fleet.  We model this task as a multiple Traveling Salesman Problem (mTSP) with $N_R$ salesmen (one per robot) sharing a common depot $\mathbf{d}$. Each swath $S_m$ is represented as a single node in an Asymmetric TSP (ATSP) cost matrix, and the LKH solver~\cite{helsgaun2017extension} is configured with a \texttt{MINSUM} objective:
\begin{equation}
\min_{\Pi_1,\dots,\Pi_{N_R}}
\sum_{r=1}^{N_R} L_r, \quad \text{s.t.} \quad |\mathcal{T}_r| \geq \left\lfloor \frac{N_S}{N_R} \right\rfloor, \;\; \forall\, r,
\label{eqr:tsp}
\end{equation}
where $L_r$ includes both the swath lengths and the transition distances for robot $r$, and $|\mathcal{T}_r|$ denotes the number of swaths assigned to robot $r$. The minimum tour size constraint, enforced via LKH's \texttt{MTSP\_MIN\_SIZE} parameter, ensures that each robot receives at least $\lfloor N_S / N_R \rfloor$ swaths, preventing degenerate allocations and promoting balanced workloads. Combined with the \texttt{MINSUM} objective, this produces spatially contiguous, approximately balanced tours that indirectly minimize the makespan.

    \subsection{Sweep Path Generation using Alternating-Edge Traversal and Visibility Graph (VG) Detouring}
\label{subsec:sweep-path}
\begin{table*}[h!]
\caption{ Comparison of all evaluated methods across multiple environments for a team of three AAVs. Reported metrics include overall computation time, total fleet energy consumption including depot round-trips ($E_t$), resulting path length ($\ell$), coverage-only fleet energy consumption ($\bar{E}_t$), and coverage-only path length ($\bar{\ell}$). The best value for each environment is highlighted in \textbf{bold}. A dash (—) indicates that no valid solution was obtained by that method in the corresponding environment.}

\label{tab:energy}
\centering
\adjustbox{max width=0.9\textwidth}
{
\large 
\renewcommand{\arraystretch}{1.15}
\begin{tabular}{l|rrrrr|rrrrr|rrrrr|rrrrr}
\hline
 & \multicolumn{5}{c|}{MRCPP (Ours)} & \multicolumn{5}{c|}{EAMCMP~\cite{datsko2024energy}} & \multicolumn{5}{c|}{DARP+MST~\cite{apostolidis2022cooperative}} & \multicolumn{5}{c}{POPCORN+SALT~\cite{shah2022large}} \\ \cline{2-21}
Scenario
  & \multicolumn{1}{c}{\begin{tabular}[c]{@{}c@{}}\rule{0pt}{2.5ex}Time \\ {[}s{]}\end{tabular}}
  & \multicolumn{1}{c}{\begin{tabular}[c]{@{}c@{}}\rule{0pt}{2.5ex}$E_t$ \\ {[}Wh{]}\end{tabular}}
  & \multicolumn{1}{c}{\begin{tabular}[c]{@{}c@{}}\rule{0pt}{2.5ex}$\ell$ \\ {[}km{]}\end{tabular}}
  & \multicolumn{1}{c}{\begin{tabular}[c]{@{}c@{}}\rule{0pt}{2.5ex}$\bar{E}_t$ \\ {[}Wh{]}\end{tabular}}
  & \multicolumn{1}{c|}{\begin{tabular}[c]{@{}c@{}}\rule{0pt}{2.5ex}$\bar{\ell}$ \\ {[}km{]}\end{tabular}}
  & \multicolumn{1}{c}{\begin{tabular}[c]{@{}c@{}}\rule{0pt}{2.5ex}Time \\ {[}s{]}\end{tabular}}
  & \multicolumn{1}{c}{\begin{tabular}[c]{@{}c@{}}\rule{0pt}{2.5ex}$E_t$ \\ {[}Wh{]}\end{tabular}}
  & \multicolumn{1}{c}{\begin{tabular}[c]{@{}c@{}}\rule{0pt}{2.5ex}$\ell$ \\ {[}km{]}\end{tabular}}
  & \multicolumn{1}{c}{\begin{tabular}[c]{@{}c@{}}\rule{0pt}{2.5ex}$\bar{E}_t$ \\ {[}Wh{]}\end{tabular}}
  & \multicolumn{1}{c|}{\begin{tabular}[c]{@{}c@{}}\rule{0pt}{2.5ex}$\bar{\ell}$ \\ {[}km{]}\end{tabular}}
  & \multicolumn{1}{c}{\begin{tabular}[c]{@{}c@{}}\rule{0pt}{2.5ex}Time \\ {[}s{]}\end{tabular}}
  & \multicolumn{1}{c}{\begin{tabular}[c]{@{}c@{}}\rule{0pt}{2.5ex}$E_t$ \\ {[}Wh{]}\end{tabular}}
  & \multicolumn{1}{c}{\begin{tabular}[c]{@{}c@{}}\rule{0pt}{2.5ex}$\ell$ \\ {[}km{]}\end{tabular}}
  & \multicolumn{1}{c}{\begin{tabular}[c]{@{}c@{}}\rule{0pt}{2.5ex}$\bar{E}_t$ \\ {[}Wh{]}\end{tabular}}
  & \multicolumn{1}{c|}{\begin{tabular}[c]{@{}c@{}}\rule{0pt}{2.5ex}$\bar{\ell}$ \\ {[}km{]}\end{tabular}}
  & \multicolumn{1}{c}{\begin{tabular}[c]{@{}c@{}}\rule{0pt}{2.5ex}Time \\ {[}s{]}\end{tabular}}
  & \multicolumn{1}{c}{\begin{tabular}[c]{@{}c@{}}\rule{0pt}{2.5ex}$E_t$ \\ {[}Wh{]}\end{tabular}}
  & \multicolumn{1}{c}{\begin{tabular}[c]{@{}c@{}}\rule{0pt}{2.5ex}$\ell$ \\ {[}km{]}\end{tabular}}
  & \multicolumn{1}{c}{\begin{tabular}[c]{@{}c@{}}\rule{0pt}{2.5ex}$\bar{E}_t$ \\ {[}Wh{]}\end{tabular}}
  & \multicolumn{1}{c}{\begin{tabular}[c]{@{}c@{}}\rule{0pt}{2.5ex}$\bar{\ell}$ \\ {[}km{]}\end{tabular}} \\ \hline
Cape
  & \textbf{1.28} & \textbf{586.37} & 36.68 & \textbf{470.08} & 29.28
  & 4.93 & 601.68 & 35.66 & 496.56 & 28.90
  & 33.82 & 600.85 & \textbf{34.92} & 503.98 & \textbf{28.68}
  & 55.87 & 818.99 & 39.16 & 708.18 & 32.06 \\
Complex-12
  & \textbf{3.60} & \textbf{478.14} & 27.40 & \textbf{455.74} & 26.12
  & — & — & — & — & —
  & 45.31 & 482.23 & \textbf{24.73} & 479.58 & \textbf{24.62}
  & — & — & — & — & — \\
Complex-22
  & \textbf{1.23} & \textbf{256.92} & 14.49 & \textbf{241.55} & 13.66
  & 1.25 & 419.69 & 23.55 & 404.95 & 22.67
  & 18.22 & 270.75 & \textbf{13.69} & 265.88 & \textbf{13.46}
  & — & — & — & — & — \\
Wetland
  & 2.10 & \textbf{293.52} & 17.06 & \textbf{274.17} & 15.82
  & \textbf{0.52} & 309.67 & 17.23 & 285.93 & 15.78
  & 44.39 & 325.10 & \textbf{15.61} & 309.93 & \textbf{14.68}
  & — & — & — & — & — \\
Island
  & \textbf{0.47} & 37.82 & 1.60 & \textbf{29.77} & 1.25
  & 0.53 & \textbf{36.80} & 1.60 & 30.25 & 1.29
  & 5.21 & 37.61 & \textbf{1.37} & 33.85 & \textbf{1.18}
  & — & — & — & — & — \\
Rect
  & 0.44 & \textbf{68.21} & \textbf{3.66} & \textbf{55.16} & \textbf{2.99}
  & \textbf{0.02} & 85.25 & 4.35 & 73.83 & 3.75
  & 4.75 & 89.18 & 3.81 & 87.37 & 3.74
  & 20.71 & 137.06 & 3.99 & 131.99 & 3.75 \\
Simple
  & 0.63 & \textbf{117.90} & 6.67 & \textbf{104.46} & \textbf{5.97}
  & \textbf{0.02} & 185.09 & 10.41 & 169.56 & 9.57
  & 5.70 & 140.25 & \textbf{6.33} & 138.14 & 6.26
  & 32.28 & 242.59 & 7.18 & 235.47 & 6.83 \\ \hline
\end{tabular}
}
\vspace{-8 pt}
\end{table*}

Once the swath assignments step is completed, we then compute continuous coverage paths using alternating-edge traversal and VG-based detouring.
The swath traversal direction and robot heading are determined \textit{post~hoc}, after the mTSP allocation, rather than being embedded as state variables in the mTSP or VG formulations.
Let $I_r = \{ k_1, k_2, \dots, k_{|I_r|} \}$ denote the swaths assigned to robot $r$. This index set represents the ordered subset of swaths allocated to robot $r$ by the mTSP solver.
To generate a continuous sweep path, we choose the waypoints based on the robot heading direction along a swath segment in such a way that produces a boustrophedon (back-and-forth) traversal along two consecutive swaths as:
\begin{equation}
(w^{(r)}_{2i-2},\, w^{(r)}_{2i-1}) =
\begin{cases}
(a_{k_i},\, b_{k_i}), & i \text{ odd}, \\
(b_{k_i},\, a_{k_i}), & i \text{ even}.
\end{cases}
\end{equation}
The left and right directions are defined relative to the rotating bounding box's minor axis $\mathbf{v}$, and the robot's heading is initialized based on the orientation of its first assigned swath. The heading alternates for every consecutive swath pair, avoiding unnecessary reorientation.
This produces the preliminary waypoint sequence
$R_r = \langle w^{(r)}_0,\, \dots,\, w^{(r)}_{2|I_r|-1} \rangle.$
The sequence $R_r$ represents the ordered set of entry and exit waypoints for robot $r$ before obstacle-aware refinement is applied.

 In the presence of obstacles or no-fly zones, swath lines may become segmented. If the mTSP solver assigns two segments of the same original swath to the same robot, the heading remains unchanged since the segments share the same orientation. Otherwise, the heading is updated via the alternating-edge rule above. By tracking the orientation of consecutive assigned swaths, we determine the source--destination pair for the VG query.
The visibility graph $G_{\mathrm{VG}} = (V_{\mathrm{VG}},\, E_{\mathrm{VG}})$ is constructed using all swath endpoints together with uniformly sampled intermediate points along the buffered obstacle and boundary polygons.  The standard VG algorithm uses only corner vertices of polygonal obstacles as intermediate nodes; however, for large obstacles, routing exclusively through corner vertices causes the detour path to deviate significantly from the original swath lines, degrading coverage performance. We therefore sample additional waypoints along obstacle edges at intervals proportional to the swath width~$w$, so that the VG can route the robot to the nearest point on the obstacle boundary relative to the interrupted swath, thereby minimizing deviation from the original swath trajectory. Edges are added between mutually visible vertex pairs that lie entirely inside the feasible region $F'$, enabling efficient computation of the shortest collision-free transition paths between disconnected swaths~\cite{choset2000coverage}. Straight-line transitions are used whenever feasible; otherwise, a VG shortest path is computed.  The heading through these detours is governed by the geometry of the sampled obstacle boundary rather than being explicitly encoded as a state or constraint in the VG.
The refined coverage plan for robot $r$ is defined as
$\Pi_r = \mathrm{VGRefine}(R_r),$
where $\Pi_r$ denotes the complete coverage path for robot $r$ obtained by replacing obstructed straight-line transitions in $R_r$ with VG-based shortest paths.
The final multirobot coverage solution is
$\{\Pi_1,\, \Pi_2,\, \dots,\, \Pi_{N_R}\},$
which represents the set of executable coverage paths for all robots in the team.

 In summary, the visibility graph remains geometrically standard in its edge construction (mutual visibility inside $F'$) and search procedure (shortest collision-free path). The two distinctions from a conventional VG are: (1)~the node set is augmented with uniformly sampled boundary points to reduce detour deviation, and (2)~the source and destination nodes fed into $\mathrm{VGRefine}$ are determined by the heading-driven alternating-edge traversal rule. Heading is therefore used exclusively to select entry/exit endpoints per swath, not as part of the VG node/edge state.

    \section{Experimental Results}

\label{sec:experiments}

To validate the effectiveness of our proposed method, we conducted both simulated and real-world experiments. We compared our approach against state-of-the-art methods in simulation environments. All benchmark scenarios are adopted from~\cite{datsko2024energy}, with the addition of a new \textit{Wetland} scenario in which AAVs are tasked with monitoring a wetland while avoiding a non-convex no-fly zone.

\begin{table}[t]
\caption{Scalability comparison in the Complex~22 and Cape environments for fleets of 4, 6, 8, and 10 AAVs. Reported metrics include computation time ($t_c$), total fleet energy consumption including depot round-trips ($E_t$), and total coverage path length ($\ell$) for MRCPP (ours) and EAMCMP~\cite{datsko2024energy}.}
\vspace{-7 pt}
\label{tab:scale}
\centering
\setlength{\tabcolsep}{3pt}
\renewcommand{\arraystretch}{0.80}
\begin{adjustbox}{width=0.85\columnwidth}
\begin{tabular}{l | r r r | r r r }
\toprule
Scenario & \multicolumn{3}{c|}{MRCPP} & \multicolumn{3}{c}{EAMCMP\cite{datsko2024energy}} \\
 & $t_c$ [s] & ${E}_t$ [Wh] & $\ell$ [km] & $t_c$ [s] & ${E}_t$ [Wh] & $\ell$ [km] \\
\midrule
complex (4 AAVs)  & \textbf{1.41} & \textbf{294.40} & \textbf{16.40} & 1.45          & 303.28 & 16.83 \\
complex (6 AAVs)  & 1.81          & \textbf{291.28} & \textbf{16.56} & \textbf{1.67} & 311.80 & 17.36 \\
complex (8 AAVs)  & 2.51          & \textbf{316.95} & \textbf{17.94} & \textbf{1.89} & 321.90 & 18.39 \\
complex (10 AAVs) & \textbf{2.90} & \textbf{334.74} & \textbf{18.91} & —             & —      & —     \\
cape (4 AAVs)     & \textbf{1.96} & \textbf{684.63} & \textbf{43.06} & 479.20        & 783.14 & 47.64 \\
cape (6 AAVs)     & \textbf{2.50} & \textbf{711.88} & \textbf{44.71} & —             & —      & —     \\
cape (8 AAVs)     & \textbf{3.15} & \textbf{809.37} & \textbf{50.95} & —             & —      & —     \\
cape (10 AAVs)    & \textbf{3.70} & \textbf{923.47} & \textbf{58.28} & —             & —      & —     \\
\bottomrule
\end{tabular}
\end{adjustbox}

\end{table}

\subsection{Benchmark}

Table~\ref{tab:energy} summarizes the benchmark results across all environments for a team of three AAVs. The swath widths are set to 31, 12, 22, 10, 8, 8, and 9\,m for the Cape, Complex-12, Complex-22, Wetland, Island, Rect, and Simple scenarios, respectively. All runs use a safe buffer equal to one swath width $w$.

\textbf{Success rate and computation time.}
MRCPP finds a valid solution for all seven environments in under 4\,s on a general-purpose computer, making it a suitable candidate for browser-based planning. In contrast, EAMCMP~\cite{datsko2024energy} fails to produce a solution for the \emph{Complex-12} environment, while POPCORN+SALT~\cite{shah2022large} fails in all environments containing obstacles or no-fly zones. DARP+MST~\cite{apostolidis2022cooperative} solves all environments but requires significantly longer computation times, particularly in the presence of obstacles. For instance, 44\,s in the \emph{Wetland} scenario and 33\,s in the \emph{Cape} scenario. It is worth noting that MRCPP is implemented using a behavior-tree library for modularity and maintainability. This design introduces a small overhead, which explains why EAMCMP~\cite{datsko2024energy} achieves marginally faster runtimes in the simpler \emph{Rect} and \emph{Simple} environments.

\textbf{Energy consumption.}
MRCPP outperforms all other planners in both total energy consumption ($E_t$) while considering user defined initial positions and total energy consumption ($E_t$) with algorithm defined initial positions inside the ROI ($\bar{E}_t$) across all environments except \emph{Island}, where EAMCMP~\cite{datsko2024energy} achieves a marginally lower total energy ($E_t = 36.80$\,Wh versus $37.82$\,Wh). This is because, unlike EAMCMP, MRCPP does not optimize coverage paths based on the initial positions of the vehicles. Nevertheless, MRCPP consistently delivers the most energy-efficient coverage plans, particularly in environments with complex regions of interest and no-fly zones. For example, in \emph{Rect}, MRCPP reduces $\bar{E}_t$ by approximately 25\% compared to EAMCMP~\cite{datsko2024energy} and by 37\% compared to DARP+MST~\cite{apostolidis2022cooperative}. These gains are attributable to MRCPP's orientation-optimized swath generation, which minimizes the number of turns and associated deceleration--acceleration energy losses, and its mTSP-based workload allocation, which ensures a uniform distribution of effort across the fleet.

\textbf{Path length.}
DARP+MST~\cite{apostolidis2022cooperative} generates the shortest paths in most environments (e.g., \emph{Complex-12}, \emph{Complex-22}, \emph{Wetland}, and \emph{Island}). However, these paths typically involve numerous directional reversals, which are inefficient from an energy standpoint---explaining why shorter geometric paths do not translate to lower energy consumption. MRCPP remains competitive in path length throughout; in \emph{Rect} and \emph{Simple}, it achieves the shortest paths outright, and in the remaining environments the differences are negligible. This suggests that MRCPP strikes a favorable balance between path length and energy efficiency.

\textbf{Scalability.}
Table~\ref{tab:scale} reports the scalability results as the number of AAVs
increases from 4 to 10 in two NFZ-constrained environments: \emph{Complex\,22}
and \emph{Cape}. MRCPP scales gracefully across all fleet sizes, with
computation times ranging from 1.41\,s (4~AAVs) to 3.70\,s (10~AAVs). In
\emph{Complex\,22}, EAMCMP~\cite{datsko2024energy} achieves marginally faster
runtimes for 6 and 8~AAVs but fails to produce a valid solution for 10~AAVs.
The gap is significantly larger in the \emph{Cape} environment, where EAMCMP
requires 479.20\,s for 4~AAVs and fails entirely for all larger fleet sizes,
whereas MRCPP solves every \emph{Cape} configuration in under 4\,s. In terms of
energy consumption, MRCPP achieves the lowest $\bar{E}_t$ for every configuration in
which both planners return a valid solution, with improvements ranging from
1.5\% to 6.6\% in \emph{Complex\,22} and 12.6\% in \emph{Cape} at 4~AAVs
(684.63\,Wh versus 783.14\,Wh). MRCPP also generates shorter coverage paths in
all comparable configurations, with reductions between 2.5\% and 9.6\%,
directly contributing to the observed energy savings through fewer traversal
segments and lower cumulative aerodynamic losses.

\subsection{Ablation Study}
 \begin{table}[t]
 \caption{Orientation ablation: fleet energy with depot round-trips ($E_t$)
  and coverage-only energy ($\bar{E}_t$), both in Wh.  Bold marks the
  lowest $E_t$ per scenario.}
  \vspace{-7 pt}
  \label{tab:orient}
  \centering
  \resizebox{\columnwidth}{!}{%
  \begin{tabular}{l rr rr rr rr}
  \toprule
   & \multicolumn{2}{c}{MAR} & \multicolumn{2}{c}{Angle Search}
   & \multicolumn{2}{c}{PCA} & \multicolumn{2}{c}{Min-Width} \\
  \cmidrule(lr){2-3}\cmidrule(lr){4-5}\cmidrule(lr){6-7}\cmidrule(lr){8-9}
  Scenario & $E_t$ & $\bar{E}_t$ & $E_t$ & $\bar{E}_t$
           & $E_t$ & $\bar{E}_t$ & $E_t$ & $\bar{E}_t$ \\
  \midrule
  Cape       & 589.21 & 464.64 & 662.84 & 537.14 & \textbf{586.37} & 470.08 & 652.80 &
  540.78 \\
  Complex-12 & 533.65 & 506.20 & 486.15 & 457.19 & 507.58 & 470.05 & \textbf{478.14} &
  455.74 \\
  Complex-22 & 275.91 & 253.37 & 278.34 & 260.82 & \textbf{256.92} & 241.55 & 276.91 &
  257.23 \\
  Island     &  39.00 &  32.43 &  40.67 &  33.13 &  39.60 &  31.57 & \textbf{37.82} &
  29.77 \\
  Rect       &  72.09 &  59.45 &  72.76 &  61.70 &  74.71 &  62.27 & \textbf{68.21} &
  55.16 \\
  Simple   & 161.45 & 143.66 & 119.85 & 104.45 & \textbf{117.90} & 104.46 & 123.98 &
  108.65 \\
    Wetland   & 302.25 & 285.79 & 302.98 & 295.38 & 293.52 & 273.76 & \textbf{289.37} &
  280.33 \\
  \bottomrule
  \end{tabular}%
  }
  
  \end{table}

\begin{figure}[t]
\centering
\subfloat[EAMCMP~\cite{datsko2024energy}\label{fig:eamcmp_solution}]{
    \includegraphics[
        width=0.47\columnwidth,
        trim=5cm 2cm 5cm 1.0cm, clip
    ]{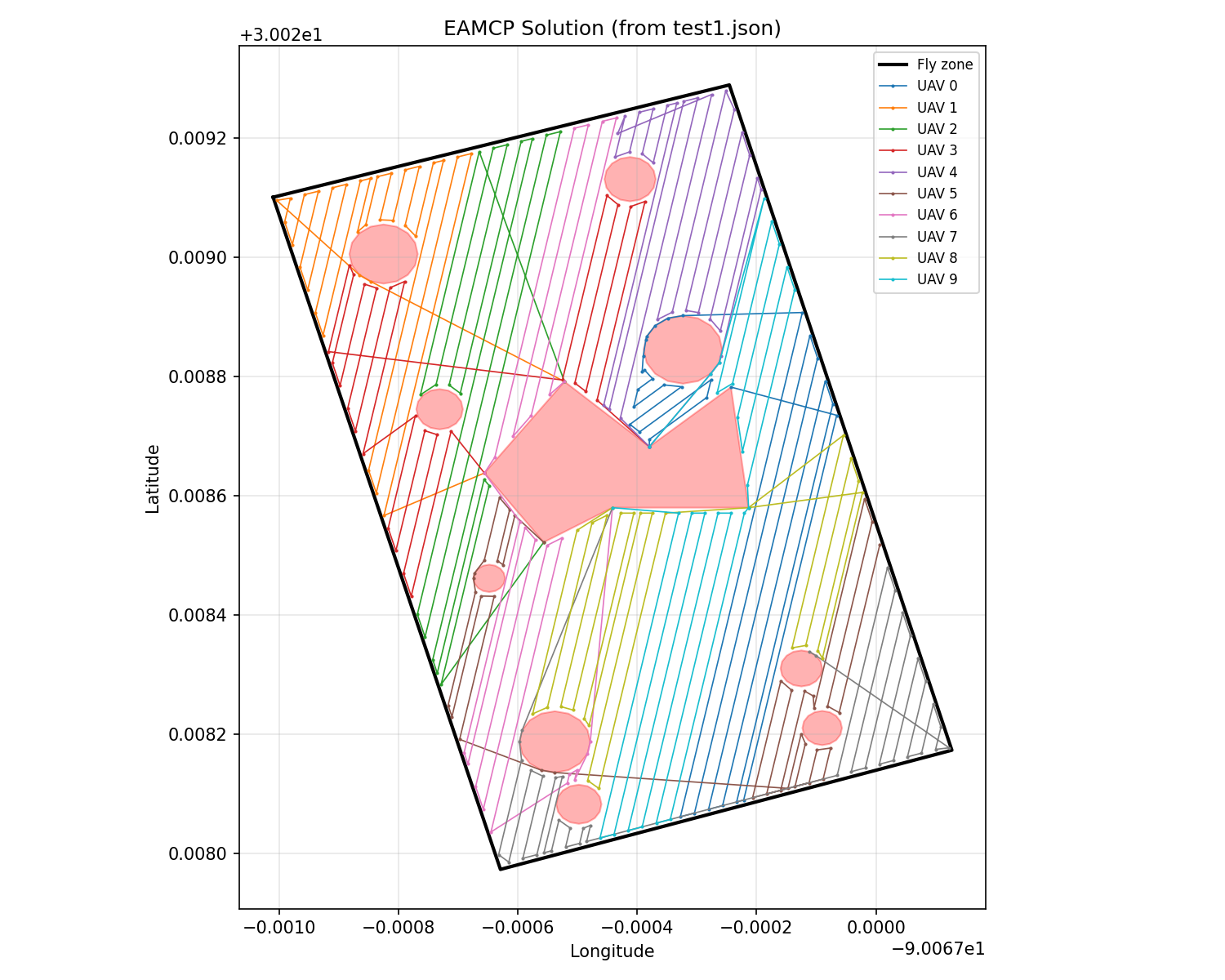}
}
\hfil
\subfloat[MRCPP\label{fig:mrcpp_solution}]{
    \includegraphics[
        width=0.47\columnwidth,
        trim=5cm 2cm 5cm 1.0cm, clip
    ]{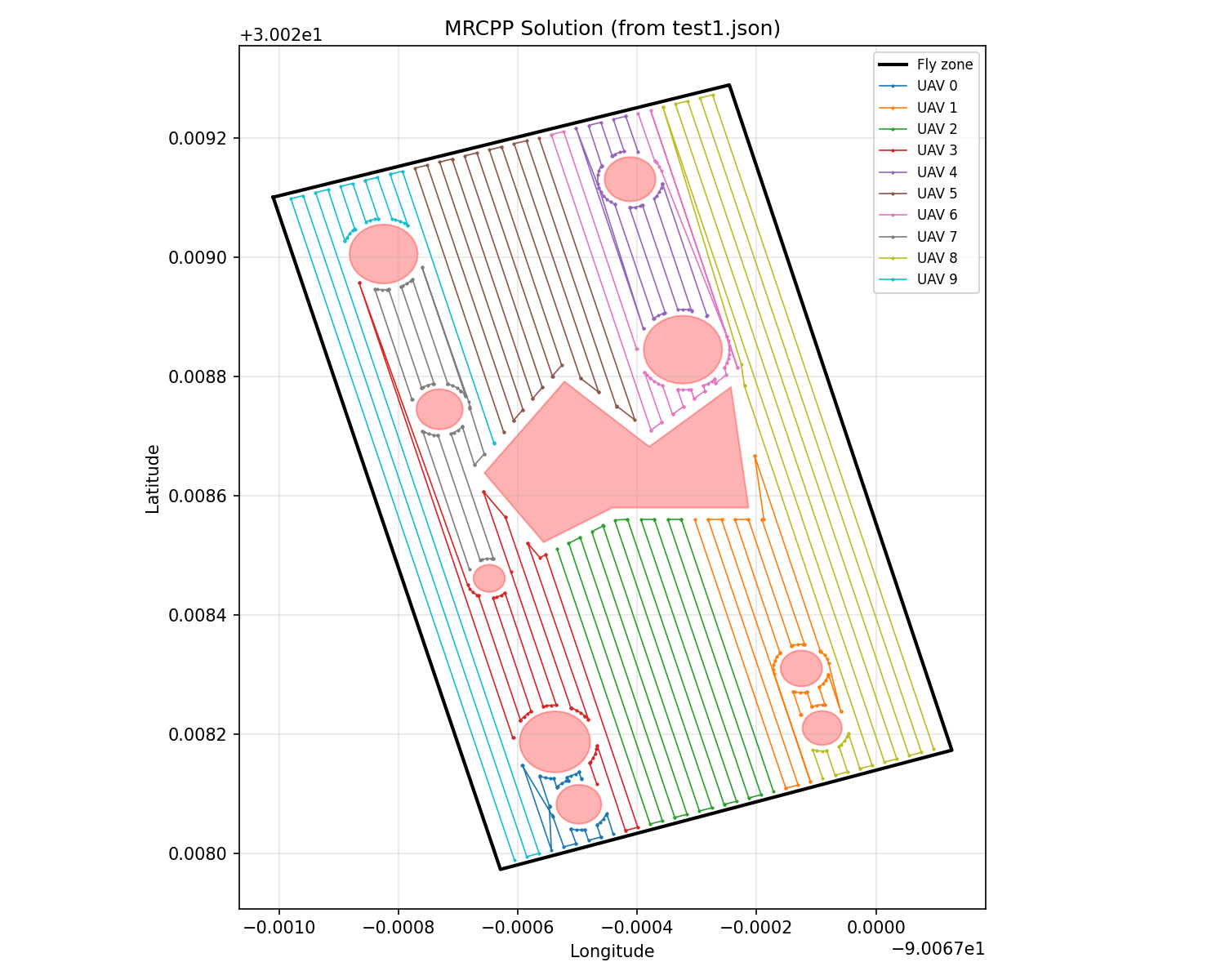}
}
\caption{Coverage paths generated by (a)~EAMCMP and (b)~MRCPP for a fleet of 10 AAVs in the presence of convex and nonconvex obstacles.}
\label{fig:obstacle_comparison}

\end{figure}

We evaluate four sweep orientation strategies: Minimum-Area Rectangle (MAR), exhaustive angle search, Principal Component Analysis (PCA), and minimum-width, across seven benchmark scenarios. The initial position is fixed at the center of the ROI and excluded from energy estimation. Consequently, robots commence task execution directly from their designated path starting points.
Table~\ref{tab:orient} reports the total fleet energy $\bar{E}_t$ (excluding initial position)  and the worst energy performance $E_{o_{\max}}$, both in Wh. Minimum-width orientation achieves the lowest $E_t$ in three of seven scenarios (Complex-12, Island, and Rect), while PCA is best in three others (Cape, Complex-22, and Simple). MAR and exhaustive angle search are never the top performers. The spread between the best and worst orientation can exceed 35\% (Simple), confirming that sweep direction is a significant design parameter for multi-robot coverage.
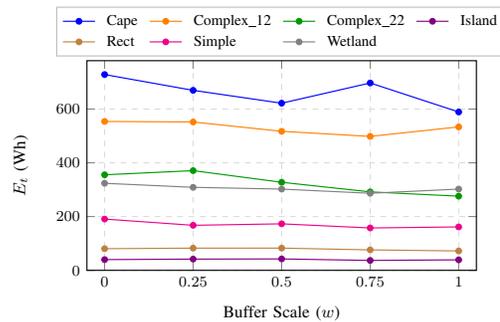
\begin{figure}[t]
\centering
\begin{adjustbox}{width=0.75\columnwidth}
\begin{tikzpicture}
\begin{axis}[
    width=\columnwidth,
    height=5.5cm,
    xlabel={Buffer Scale ($w$)},
    ylabel={$E_t$ (Wh)},
    grid=both,
    grid style={dashed, gray!30},
    legend style={
        at={(0.5,1.02)},
        anchor=south,
        font=\footnotesize,
        cells={anchor=west},
        draw=gray!50,
        legend columns=4,
        column sep=4pt,
    },
    legend cell align={left},
    mark size=1.5pt,
    semithick,
    xmin=-0.05, xmax=1.05,
    ymin=0, ymax=780,
    xtick={0, 0.25, 0.5, 0.75, 1.0},
    label style={font=\small},
    tick label style={font=\footnotesize},
]

\addplot[color=blue, mark=*] coordinates {
    (0.0, 728.5099)
    (0.25, 669.7216)
    (0.5, 621.8866)
    (0.75, 697.1313)
    (1.0, 589.2081)
};
\addlegendentry{Cape}

\addplot[color=orange, mark=*] coordinates {
    (0.0, 553.9357)
    (0.25, 551.9174)
    (0.5, 517.2515)
    (0.75, 498.2825)
    (1.0, 533.648)
};
\addlegendentry{Complex\_12}

\addplot[color=green!60!black, mark=*] coordinates {
    (0.0, 355.4124)
    (0.25, 370.8885)
    (0.5, 327.6012)
    (0.75, 291.8563)
    (1.0, 275.9082)
};
\addlegendentry{Complex\_22}

\addplot[color=violet, mark=*] coordinates {
    (0.0, 39.8488)
    (0.25, 41.7244)
    (0.5, 42.3898)
    (0.75, 36.7398)
    (1.0, 39.0039)
};
\addlegendentry{Island}

\addplot[color=brown, mark=*] coordinates {
    (0.0, 80.4444)
    (0.25, 82.5099)
    (0.5, 82.6215)
    (0.75, 76.0233)
    (1.0, 72.0886)
};
\addlegendentry{Rect}

\addplot[color=magenta, mark=*] coordinates {
    (0.0, 190.6886)
    (0.25, 167.4918)
    (0.5, 172.9453)
    (0.75, 157.5903)
    (1.0, 161.4503)
};
\addlegendentry{Simple}

\addplot[color=gray, mark=*] coordinates {
    (0.0, 323.7851)
    (0.25, 308.799)
    (0.5, 302.2236)
    (0.75, 286.668)
    (1.0, 302.2547)
};
\addlegendentry{Wetland}

\end{axis}
\end{tikzpicture}
\end{adjustbox}
\vspace{-10 pt}
\caption{Effect of buffer scale $w$ on the total fleet energy consumption $E_t$ across different environments. Lower values indicate better energy efficiency.}
\label{fig:buffer_scale_energy}
\end{figure}

\begin{table}[t]
\caption{Performance comparison of MRCPP and EAMCMP across varying fleet sizes in the presence of 10 convex and nonconvex obstacles. Bold entries indicate the best value in each metric per fleet configuration.}
\vspace{-7 pt}
\label{tab:obstacle_comparison}
\centering
\footnotesize
\renewcommand{\arraystretch}{0.75}
\setlength{\tabcolsep}{4pt}
\sisetup{round-mode=places, round-precision=2, table-format=3.4}
\begin{tabular}{@{}cl
    S[table-format=2.4]
    S[table-format=3.4]
    S[table-format=2.2]@{}}
\toprule
{\#Robots} & {Method} & {$E_{o_{\max}}$ (Wh)} & {$\bar{E}_t$ (Wh)} & {$T_c$ (s)} \\
\midrule
\multirow{2}{*}{3}
  & MRCPP  & \textbf{41.8050} & \textbf{111.1718} & \textbf{9.65}  \\
  & EAMCMP~\cite{datsko2024energy} & 44.4855           & 132.2499           & 68.87          \\
\midrule
\multirow{2}{*}{6}
  & MRCPP  & \textbf{21.8133} & \textbf{104.5834} & \textbf{3.22}  \\
  & EAMCMP~\cite{datsko2024energy} & 23.1114           & 135.5108           & 58.06          \\
\midrule
\multirow{2}{*}{10}
  & MRCPP  & \textbf{13.9163} & \textbf{96.7328}  & \textbf{3.11}  \\
  & EAMCMP~\cite{datsko2024energy} & 14.7268           & 139.7665           & 68.95          \\
\bottomrule
\end{tabular}

\end{table}

Table~\ref{tab:obstacle_comparison} compares MRCPP and EAMCMP~\cite{datsko2024energy} on the obstacle-rich environment depicted in Fig.~\ref{fig:envs}, which contains 10 convex and nonconvex obstacles, for fleet sizes of 3, 6, and 10 AAVs.
\emph{Computational efficiency:} The average computation time for MRCPP is $5.33$\,s compared to $65.29$\,s for EAMCMP, making MRCPP roughly 12 times faster on average. Furthermore, EAMCMP's computation time remains relatively flat ($\approx$58--69\,s) regardless of fleet size, whereas MRCPP scales more efficiently, with computation time decreasing from $9.65$\,s at 3 AAVs to $3.11$\,s at 10 AAVs.
\emph{Energy efficiency:} MRCPP demonstrates superior scalability, with total fleet energy consumption decreasing by $14.44$\,Wh ($\approx$13.0\%) as the fleet grows from 3 to 10 AAVs, reflecting effective workload distribution through its balanced mTSP formulation and efficient path handling via the proposed VG detour algorithm. Conversely, EAMCMP's total energy consumption increases by $7.52$\,Wh ($\approx$5.7\%) over the same range, as additional AAVs introduce redundant transitions due to its sensitivity to initial positions rather than the geometric structure of the coverage area.
\emph{Safety and feasibility:} While EAMCMP occasionally violates NFZ and obstacle constraints near hazards, as shown in Fig.~\ref{fig:eamcmp_solution}, MRCPP maintains safety through explicit buffer-zone constraints. By integrating intermediate points via the VG detour algorithm, MRCPP ensures a strictly enforced minimum separation distance between the planned paths and obstacle boundaries as shown in Fig.~\ref{fig:mrcpp_solution}.

Fig.~\ref{fig:buffer_scale_energy} shows the sensitivity of irregular geometries to the buffer parameter. In non-convex environments, energy consumption fluctuates more noticeably as the buffer parameter increases. For instance, in Complex-12, energy is minimized at $0.75w$ (498.28\,Wh) since a larger buffer over-shrinks narrow corridors near the NFZ, whereas for Cape, energy is minimized at $1.0w$ (589.21\,Wh) since the full buffer prunes fragmented swaths in narrow coastal protrusions that otherwise cause expensive transitions. In contrast, simpler environments such as Island, Rect, Simple, and Wetland exhibit a nearly flat response. 

\subsection{Real World Experiment}

We conducted two categories of real-world experiments to validate the proposed framework: aerial coverage using AAVs and surface coverage using ASVs.
For the aerial experiments, we deployed two DJI Mavic Air AAVs to cover a polygonal ROI. We uploaded the planned coverage paths to the Litchi mission-planning software, which then executed each trajectory autonomously on the corresponding AAV. The two AAVs together covered 615 m, with AAV 1 flying 306 m and AAV 2 flying 309 m. Because we deployed both vehicles concurrently, they completed the mission in 78 s. Their average flight speeds were 4.25 m/s for AAV 1 and 4.41 m/s for AAV 2.
As shown  on the left side of Fig.~\ref{fig:field_exp}, the executed paths appear in orange and blue, and the circular no-fly zone (NFZ) appears in red. Some segments of the blue and orange paths are not perfectly straight because wind disturbances affected the AAVs during flight.
We further evaluated the proposed method in a coastal bay environment using two ASVs, each equipped with a YSI EXO2 sonde for in situ water-quality sensing. The ASVs were assigned to cover a designated ROI while avoiding a yellow polygonal obstacle analogous to the aerial experiments. As shown on right side of Fig.~\ref{fig:field_exp}, the left panel presents a live video feed from the ASVs, while the right panel displays their real-time positions and executed coverage paths overlaid on satellite imagery. This interface facilitated real-time monitoring and verification of coverage execution throughout the mission. The path length for ASV1 was 138.15 m (marked in red), and the path length for ASV2 was 125.71 m (marked in green), as shown in the right panel.

    \section{Conclusion}
    This letter introduces an energy efficient multi-robot coverage path planning framework for complex environments with obstacles and no-fly zones. The proposed framework achieves this through minimum-turn swath generation, workload-balanced task assignment, and obstacle and boundary-aware sweep path generation. The framework is evaluated on several key metrics, including computation time, path length, and energy consumption (both trajectory-based and trajectory-free). Extensive simulations demonstrated that the proposed method outperformed state-of-the-art planners by an order of magnitude. The framework is validated in experiments using multiple AAVs and ASVs. These experiments confirm the framework's capability to generate reliable, well-balanced coverage paths under realistic conditions. Future work will include trajectory planning to efficiently avoid dynamic obstacles during full coverage in complex environments.

    \bibliography{main}
    \bibliographystyle{IEEEtran}

\end{document}